\documentclass{article}
\usepackage{spconf,amsmath,graphicx}
\usepackage{amssymb}
\usepackage{pifont}%
\usepackage{multirow}
\usepackage{subcaption}

\title{Semi-supervised learning via Feedforward-Designed \\
Convolutional Neural Networks}

\name{Yueru Chen, Yijing Yang, Min Zhang and C.-C. Jay Kuo}
\address{University of Southern California, Los Angeles, California, USA}

\begin{document}
\ninept
\maketitle
\begin{abstract}
A semi-supervised learning framework using the feedforward-designed
convolutional neural networks (FF-CNNs) is proposed for image
classification in this work.  One unique property of FF-CNNs is that no
backpropagation is used in model parameters determination.  Since
unlabeled data may not always enhance semi-supervised learning
\cite{zhu2006semi}, we define an effective quality score and use it to
select a subset of unlabeled data in the training process.  We conduct
experiments on the MNIST, SVHN, and CIFAR-10 datasets, and show that the
proposed semi-supervised FF-CNN solution outperforms the CNN trained
by backpropagation (BP-CNN) when the amount of labeled data is reduced.
Furthermore, we develop an ensemble system that combines the output
decision vectors of different semi-supervised FF-CNNs to boost
classification accuracy. The ensemble systems can achieve further
performance gains on all three benchmarking datasets. 
\end{abstract}

\begin{keywords}
Semi-supervised learning, Ensemble, Image classification, Interpretable CNN
\end{keywords}

\section{Introduction}\label{sec:intro}

When there are no sufficient labeled data available, we need to resort
to semi-supervised learning (SSL) in tackling machine learning problems.
SSL is essential in real world applications, where labeling is expensive
and time-consuming. It attempts to boost the performance by leveraging
unlabeled data.  Its main challenge is how to use unlabeled data
effectively to enhance decision boundaries that have been obtained by a
small amount of labeled data.  Convolutional neural networks (CNNs) are
typically a fully-supervised learning tool since they demand a large
amount of labeled data to represent the cost function accurately. As the
amount of labeled data is small, the cost function is less accurate and,
thus, it is not easy to formulate an SSL framework using CNNs. 

Model parameters of CNNs are traditionally trained under a certain
optimization framework with the backpropagation (BP) algorithm. They are
called BP-CNNs. Recently, a feedforward-designed convolutional neural
network (FF-CNN) methodology was proposed by Kuo {\em et al.}
\cite{kuo2018interpretable}.  The model parameters of FF-CNNs at a
target layer are determined by statistics of the output from its
previous layer. Neither an optimization framework nor the BP training
algorithm is utilized.  Clearly, FF-CNNs are less dependent on data
labels. We propose an SSL system based on FF-CNNs in this work. 

This work has several novel contributions. First, we apply FF-CNNs to
the SSL context and show that FF-CNNs outperforms BP-CNNs when the size
of the labeled data set becomes smaller. Second, we propose an ensemble
system that fuses the output decision vectors of multiple FF-CNNs so as
to achieve even better performance. Third, we conduct experiments in
three benchmarking datasets (i.e., MNIST, SVHN, and CIFAR-10) to
demonstrate the effectiveness of the proposed solutions as described
above. 

The rest of this work is organized as follows. Both SSL and FF-CNN are
reviewed in Sec. \ref{sec:review}.  The proposed SSL solutions using a
single FF-CNN and ensembles of multiple FF-CNNs are described in Sec.
\ref{sec:proposed}.  Experimental results are shown in
Sec.\ref{sec:experiment}. Finally, concluding remarks are drawn and
future work is discussed in Sec. \ref{sec:conclusion}. 

\section{Review of Related Work}\label{sec:review}

{\bf Semi-Supervised Learning (SSL).} There are several well-known SSL
methods proposed in the literature. Iterative learning, including
self-training \cite{rosenberg2005semi} and co-training
\cite{blum1998combining}, learns from unlabeled data that have high
confidence predictions.  Transductive SVMs
\cite{joachims1999transductive} extend standard SVMs by maximizing the
margin on unlabeled data as well. Another SSL method is to construct a
graph to represent data structures and propagate the label information
of a labeled data point to unlabeled ones \cite{zhu2002learning,
blum2004semi}.  More recently, several SSL methods are proposed based on
deep generative models, such as the variational auto-encoder (VAE)
\cite{kingma2014semi}, and generative adversarial networks (GAN)
\cite{dai2017good, salimans2016improved}.  All parameters in these
networks are determined by the stochastic gradient descent (SGD)
algorithm through BP and they are trained based on both labeled and
unlabeled data. 

{\bf Feedforward-designed CNNs (FF-CNNs).} The BP training is
computationally intensive, and the learning model of a BP-CNN is lack of
interpretability. New solutions have been proposed to tackle these
issues.  Examples include: interpretable CNNs
\cite{zhang2017interpretable, kuo2016understanding, kuo2017cnn} and
feedforward-designed CNNs (FF-CNNs) \cite{chen2017saak, kuo2018data,
kuo2018interpretable}.  FF-CNNs contain two modules: 1) construction of
convolutional (conv) layers through subspace approximations, and 2)
construction of fully-connected (FC) layers via training sample
clustering and least-squared regression (LSR). They are elaborated
below. 

The construction of conv layers is realized by multi-stage Saab
transforms \cite{kuo2018interpretable}.  The Saab transform is a variant
of the principal component analysis (PCA) with a constant bias vector to
annihilate activation's nonlinearity. The Saab transform can reduce
feature redundancy in the spectral domain, yet there still exists
correlation among spatial dimensions of the same spectral component.
This is especially true in low-frequency spectral components. Thus, a
channel-wise PCA (C-PCA) was proposed in \cite{chen2019ensembles} to
reduce spatial redundancy of Saab coefficients furthermore.  Since the
construction of conv layers is unsupervised, they can be fully adopted
in an SSL system. 

The construction of FC layers is achieved by the cascade of multiple
rectified linear least-squared regressors (LSRs)
\cite{kuo2018interpretable}.  Let the input and output dimensions of a
FC layer be $N_{in}$ and $N_{out}$ (with $N_{in} > N_{out}$),
respectively. To construct an FC layer, we cluster input samples of
dimension $N_{in}$ into $N_{out}$ clusters, and assign pseudo-labels
based on clustering results.  Next, all samples are transformed into a
vector space of dimension $N_{out}$ via LSR, where the index of the
output space dimension defines a pseudo-label. In this way, we obtain a
supervised LSR building module to serve as one FC layer.  It
accommodates intra-class variability while enhancing discriminability
gradually. 

\section{Proposed Methods}\label{sec:proposed}


\subsection{Semi-supervised Learning System}\label{subsec:semi}

{\bf Problem Formulation and Data Pre-processing.} The semi-supervised
classification problem can be defined as follows. We have a set of $M$
unlabeled samples
$$
{\bf X}^{ul} = \{{\bf x}^{ul}_1, \cdots , {\bf x}^{ul}_M \}, 
$$
where ${\bf x}^{ul}_i \in \mathbb{R}^{D_{in}}$ is the $i$th input unlabeled 
sample, and a set of labeled samples that can be written in form of pairs:
$$
({\bf X}^l, {\bf Y}^{l}) = \{({\bf x}^l_1, y^l_1), \cdots ,({\bf x}^l_N , y^l_N \}, 
$$
where ${\bf x}^l_i \in \mathbb{R}^{D_{in}}$ is the $i$th input labeled
sample and $y_i^{l} \in \{ 1, \cdots , L \}$ is its class label. We omit
index $i$ whenever the context is clear. 

We adopt the multi-stage Saab transforms and C-PCA for unsupervised image 
feature extraction. They can be expressed as
$$
{\bf z}^{l}=T_{saab}({\bf x}^{l}), \mbox{  and  } 
{\bf z}^{ul}=T_{saab}({\bf x}^{ul}),
$$
where ${\bf z}^{l}, {\bf z}^{ul} \in \mathbb{R}^{D_{out}}$, and
$D_{in} > D_{out}$. This is used to facilitate classification with more
powerful image features in a lower dimensional
space. 

\begin{figure}[ht!]
\centering
\centerline{\includegraphics[width=0.65\linewidth]{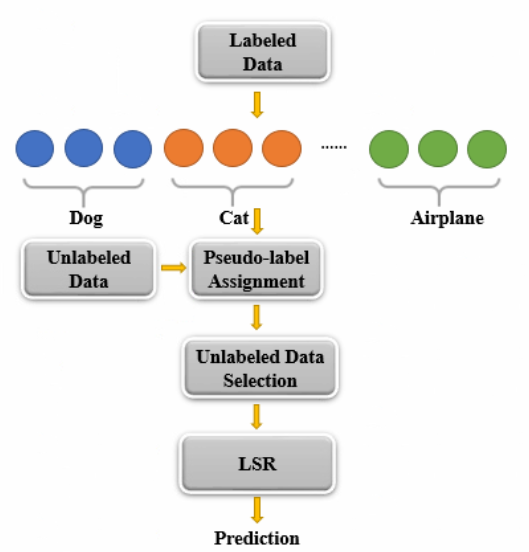}}
\caption{{The proposed semi-supervised learning (SSL) system.}\label{fig:overview-1}}
\end{figure}

{\bf Pseudo-Label Assignment.} Next, we propose a semi-supervised method
in the design of FC layers using multi-stage rectified LSRs as
illustrated in Fig. \ref{fig:overview-1}.  The pseudo-labels should be
generated for both labeled and unlabeled samples in solving the LSR
problem. To achieve this goal, we conduct K-means clustering on labeled
samples of the same class. For example, we can cluster samples of a
single original class into $M$ sub-classes to generate $M$
pseudo-categories. If there are $N$ original classes, we will generate
$M \times N$ pseudo-categories in total.  Then, pseudo-labels of
labeled data can be generated by representing pseudo categories with
one-hot vectors, denoted as 
$$
{\bf Y}^{p} = \{{\bf y}^{p}_1, \cdots, {\bf y}^{p}_M \}.
$$
The centroid of the $j$th pseudo-category, denoted by $c_j$, provides a
representative sample for the corresponding original class.  

We define the probability vector of the pseudo-category for each
unlabeled sample as
\begin{equation}\label{equ:pv}
p(t_k|{\bf z}^{ul}) = \frac{\alpha e^{{d_k}}}{\sum_{j}{\alpha e^{{d_j}}}};
\end{equation} 
where $t_k$ represents the $k$th pseudo-category and
\begin{equation}
d_k = \frac{{{\bf z}^{ul}} \cdot {\bf c}_k}{\|{\bf z}^{ul}\|\|{\bf c}_k\|}.
\end{equation}
Then, the probability vector in (\ref{equ:pv}) can be used as the
pseudo-label of each unlabeled sample to set up a system of linear
regression equations  that relates the input data samples and pseudo-labels,
$$
[{\bf Y}^{p},{\bf P}({\bf T}|{\bf Z}^{ul})] = {\bf W}_{fc}[{\bf Z}^{l},{\bf Z}^{ul}],
$$
where the parameters of FC layer are denoted by
\begin{equation}
{\bf W}_{fc} = {\bf (Z^TZ)^{-1}Z^TY}, 
\end{equation}
and where
\begin{equation}
{\bf Z} = [{\bf Z}^{l}, {\bf Z}^{ul}], \mbox{  and  } 
{\bf Y} = [{\bf Y}^{p}, {\bf P}( {\bf T}|{\bf Z}^{ul})].
\end{equation}
The final output of one-stage rectified LSR is
\begin{equation}
{\bf z}^{l}_{out} = f({\bf W}_{fc} {\bf z}^{l}) \mbox{  and  } 
{\bf z}^{ul}_{out} = f({\bf W}_{fc} {\bf z}^{ul}),
\end{equation}
where $f(.)$ is a non-linear activation function (e.g. ReLU in our
experiments), and ${\bf z}^{l}_{out}$ and ${\bf z}^{ul}_{out}$ denote
outputs of labeled and unlabeled data, respectively.  The output vectors
lie in a lower dimensional space. They are used as features to the next
stage rectified LSR. 

{\bf Unlabeled Sample Selection.} Not every unlabeled sample is suitable
for constructing FC layers. We define a quality score for each unlabeled
sample ${\bf z}^{ul}$ as
\begin{equation}\label{equ:qs}
S_{i} ({\bf z}^{ul}) = \frac{\sum_{k \in C_i}{p(t_k|{\bf z}^{ul})}}
{\sum_{j}{p(t_j|{\bf z}^{ul})}},
\end{equation} 
where $C_i$ indicates a set of pseudo-categories that belong to the
original $i$th class. A low qualify score indicates that the sample
is far away from the representative set of examples of a single original
class. We exclude those samples in solving the LSR problems. 

{\bf Multi-stage LSRs.} We repeat several LSRs and finally provide the
predicted class labels. In the last stage, we cluster input data based
on their original class labels and the LSR is solved using labeled
samples only. The multi-stage setting is needed to remove feature
redundancy in the spectral dimension and resolve intra-class variability
gradually. 

\subsection{Ensembles}\label{subsec:eb}

\begin{figure}[ht!]
\centering
\centerline{\includegraphics[width=0.65\linewidth]{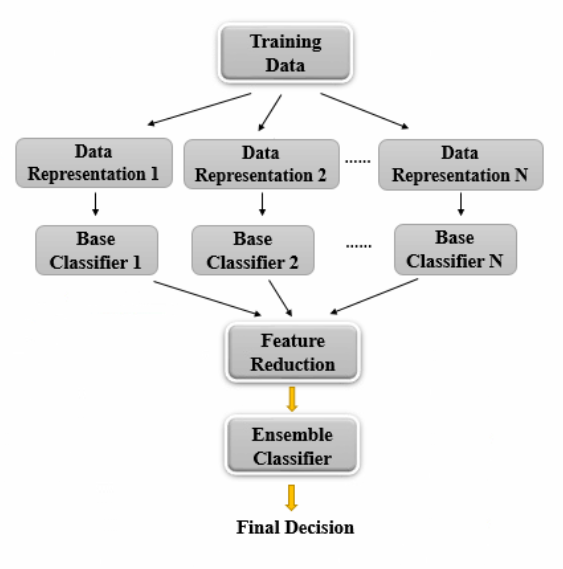}}
\caption{{An ensemble of multiple SSL decision systems.}\label{fig:overview-2}}
\end{figure}

Ensembles are often used to combine results from multiple weak
classifiers to yield a stronger one \cite{zhang2012ensemble}. We use the
ensemble idea to achieve better performance in semi-supervised
classification.  Although both BP-CNNs and FF-CNNs can be improved by
ensemble methods, FF-CNNs have much lower training complexity to justify
an ensemble solution.  The proposed ensemble system is illustrated in
Fig.  \ref{fig:overview-2}. Multiple semi-supervised FF-CNNs are adopted
as the first-stage base classifiers in an ensemble system. Their output
decision vectors are concatenated as new features. Afterwards, we apply
the principal component analysis (PCA) technique to reduce feature
dimension and then feed the dimension-reduced feature vector to the
second-stage ensemble classifier. 

High input diversity is essential to an effective ensemble system that
can reach a higher performance gain \cite{brown2005diversity,
kuncheva2003measures}. In the proposed ensemble system, we adopt three
strategies to increase input diversity. First, we consider different
filter sizes in the conv layers as illustrated in Table \ref{table:set}
since different filter sizes lead to different features at the output of
the conv layer with different receptive fields. Second, we represent
color images in different color spaces \cite{ibraheem2012understanding}.
In the experiments, we choose the RGB, YCbCr and Lab color spaces and
process three channels separately for the latter two color spaces.
Third, we decompose images into a set of feature maps using the 3x3 Laws
filters \cite{laws1980rapid}, where each feature map focuses on
different characteristics of the input image (e.g., brightness,
edginess, etc.)

\begin{table}[htb]
\begin{center}
\caption{Network architectures with respect to different input types and
different conv layer parameter settings. The second to fourth columns
indicate inputs from MNIST, RGB inputs and the single channel inputs
from SVHN and CIFAR-10, accordingly.}\label{table:set}
\begin{tabular}{|c|c|c|c|c|} \hline
\multicolumn{2}{|c|}{}  & Greyscale  & RGB & Single Channel\\ \hline
\multirow{ 2}{*}{FF-1} 
&Conv1 & 5$\times$5$\times$1, 6 & 5$\times$5$\times$3, 32 &5$\times$5$\times$1, 16 \\ 
&Conv2  & 5$\times$5$\times$6, 16 & 5$\times$5$\times$32, 64 &5$\times$5$\times$16, 32 \\ \hline
\multirow{ 2}{*}{FF-2} 
&Conv1 & 3$\times$3$\times$1, 6 & 3$\times$3$\times$3, 24 &3$\times$3$\times$1, 8 \\ 
&Conv2  & 5$\times$5$\times$6, 16 & 5$\times$5$\times$24, 64 &5$\times$5$\times$8, 32 \\ \hline
\multirow{ 2}{*}{FF-3} 
&Conv1 & 5$\times$5$\times$1, 6 & 5$\times$5$\times$3, 32 &5$\times$5$\times$1, 16 \\ 
&Conv2  & 3$\times$3$\times$6, 16 & 3$\times$3$\times$32, 64 &3$\times$3$\times$16, 32 \\ \hline
\multirow{ 2}{*}{FF-4} 
&Conv1 & 3$\times$3$\times$1, 6 & 3$\times$3$\times$3, 24 &3$\times$3$\times$1, 8 \\ 
&Conv2  & 3$\times$3$\times$6, 16 & 3$\times$3$\times$24, 48 &3$\times$3$\times$8, 24 \\ \hline
\end{tabular}
\end{center}
\end{table}

\section{Experimental Results}\label{sec:experiment}

We conduct experiments on three popular datasets: MNIST
\cite{lecun1998gradient}, SVHN \cite{netzer2011reading} and CIFAR-10
\cite{krizhevsky2009learning}. We randomly select a subset of labeled
training data from them, and test the classification performance on the
entire testing set.  Each object class has the same number of labeled
data to ensure balanced training. We adopt the LeNet-5 architecture
\cite{Lecun98gradient-basedlearning} for the MNIST dataset.  Since
CIFAR-10 is a color image dataset, we increase the filter numbers of the
first and the second conv layers and the first and the second FC layers
to 32, 64, 200 and 100, respectively, by following
\cite{kuo2018interpretable}. The C-PCA is applied to the output of the
second conv layer and the feature dimension per channel is reduced from
25 to 20 (for MNIST) or 15 (for SVHN) or 12 (for CIFAR-10).  The
probability vectors are computed using Eq. (\ref{equ:pv}), where
$\alpha$ is set to 50 for all three datasets. 

The Radial Basis Function (RBF) SVM classifier is used as the
second-stage classifier in the ensemble systems in all experiments.
Before training the SVM classifier, PCA is applied to the cascaded
decision vectors of first-stage classifiers. The reduced feature
dimension is determined based on the correlation of decision vectors of
base classifiers in an ensemble. 

\subsection{Individual Semi-Supervised FF-CNN}

\begin{figure*}[ht!]
\begin{minipage}[b]{0.33\linewidth}
  \centering
  \centerline{\includegraphics[width=0.97\linewidth]{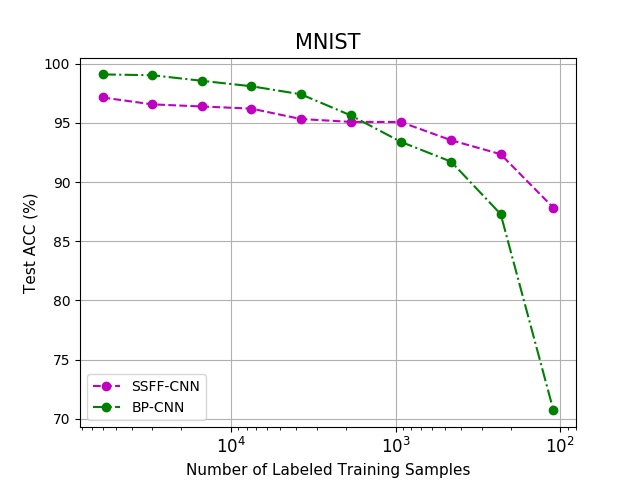}}
\end{minipage} 
\hfill
\begin{minipage}[b]{0.33\linewidth}
  \centering
  \centerline{\includegraphics[width=0.97\linewidth]{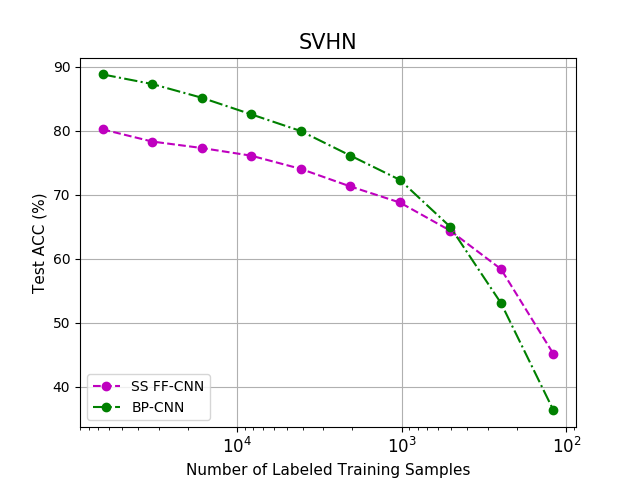}}
\end{minipage}
\hfill
\begin{minipage}[b]{0.33\linewidth}
  \centering
  \centerline{\includegraphics[width=0.97\linewidth]{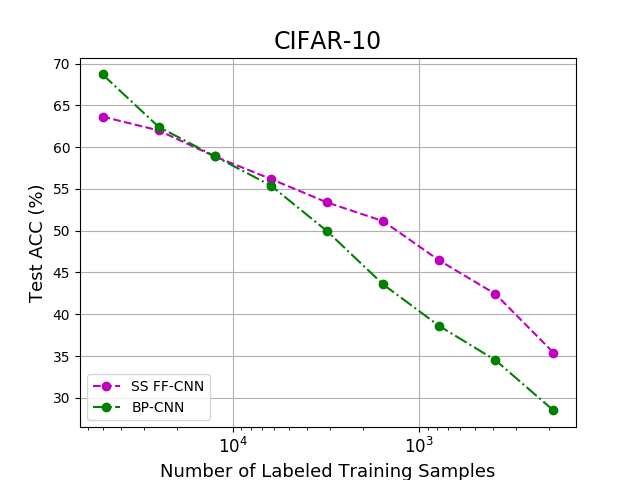}}
\end{minipage}
\caption{The comparisons of testing accuracy (\%) using BP-CNNs, and
semi-supervised FF-CNNs on MNIST, SVHN and CIFAR-10 datasets,
respectively.} \label{fig:acc_1}
\end{figure*}

We compare the performance of the BP-CNN and the proposed
semi-supervised FF-CNN on three benchmark datasets in Fig.
\ref{fig:acc_1}. For MNIST and SVHN, 1/2, 1/4, 1/8, 1/16,1/32, 1/64,
1/128, 1/256, 1/512 of the entire labeled training set are randomly
selected to train the networks.  As to CIFAR-10, 1/2, 1/4, 1/8,
1/16,1/32, 1/64, 1/128, 1/256 of the whole labeled dataset are used to
learn network parameters. We use the labeled data to train the BP-CNNs.
We select unlabeled training data with quality scores of top 70\%, 70\%
and 80\% in MNIST, SVHN and CIFAR-10, respectively, in the training of
the corresponding semi-supervised FF-CNN. 

\begin{table}[h!]
\normalsize
\centering
\caption{Testing accuracy (\%) comparison under three settings: 1)
without using unlabeled data; 2) using the entire set of unlabeled data;
and 3) using a subset of unlabeled data based on quality scores defined
by Eq. (\ref{equ:pv}).  1/256 of the labeled data is used on MNIST and
SVHN datasets, and 1/128 of the labeled data is used on CIFAR-10
dataset. }\label{table:accuracy_ab}
\vspace{0.3cm}
\begin{tabular}{|c|ccc|} \hline
            & MNIST & SVHN & CIFAR-10 \\ \hline
Setting 1 & 57.19 ($\pm$ 3.4) & 25.17 ($\pm$ 1.10) & 24.64 ($\pm$ 0.25)  \\ \hline
Setting 2 & 92.26 ($\pm$ 0.33) & 53.76 ($\pm$ 0.97) & 41.95 ($\pm$ 0.56) \\ \hline
Setting 3 & \bf 92.65 ($\pm$ 0.14) & \bf 58.58 ($\pm$ 0.78) & \bf 42.53 ($\pm$ 0.57) \\ \hline
\end{tabular}
\end{table}

\begin{figure*}[ht!]
\begin{minipage}[b]{0.33\linewidth}
  \centering
  \centerline{\includegraphics[width=0.97\linewidth]{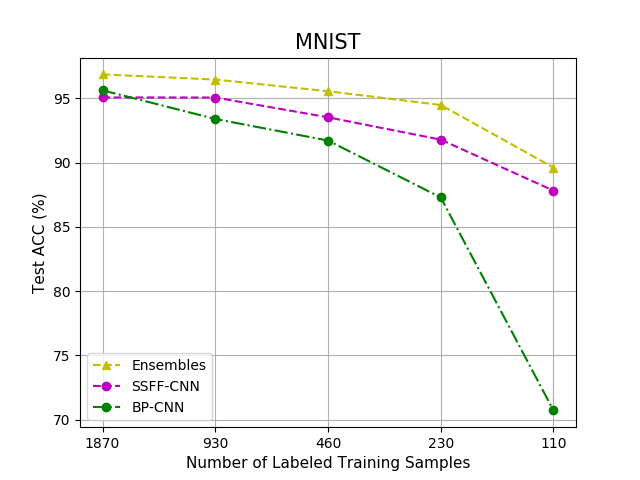}}
\end{minipage} 
\hfill
\begin{minipage}[b]{0.33\linewidth}
  \centering
  \centerline{\includegraphics[width=0.97\linewidth]{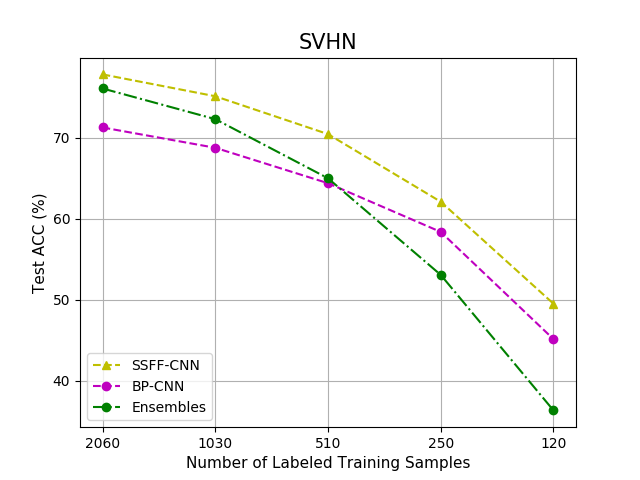}}
\end{minipage}
\hfill
\begin{minipage}[b]{0.33\linewidth}
  \centering
  \centerline{\includegraphics[width=0.97\linewidth]{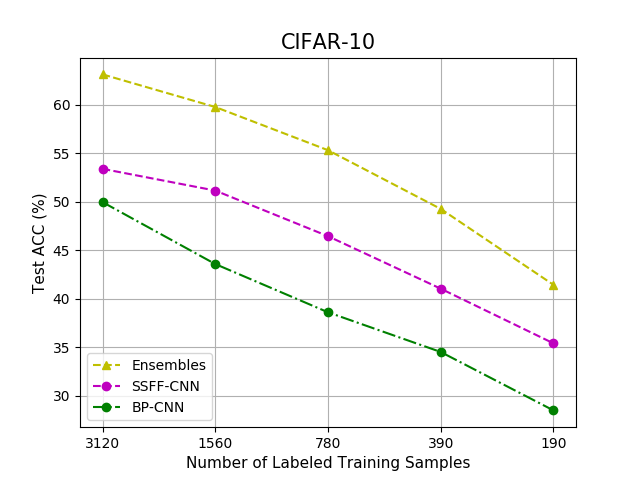}}
\end{minipage}
\caption{The comparisons of testing accuracy (\%) using BP-CNNs,
semi-supervised FF-CNNs, and ensembles of semi-supervised FF-CNNs on the
small labeled portion against MNIST, SVHN and CIFAR-10 datasets,
respectively.} \label{fig:acc_2}
\end{figure*}

\begin{figure}[ht!]
\centering
\includegraphics[width=0.8\linewidth]{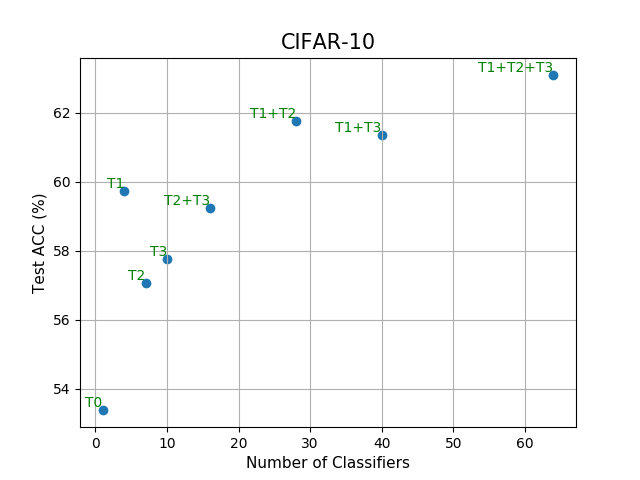}
\caption{The relation between test accuracy (\%) and the number of
semi-supervised FF-CNNs in the ensemble, where three diversity types
are indicated as T1, T2 and T3, and T0 indicates the individual
semi-supervised FF-CNN. The experiments are conducted on CIFAR-10 with
1/16 of the whole labels.} \label{fig:acc_overall}
\end{figure}

When using the entire labeled training set, the semi-supervised FF-CNN
is exactly the same as the FF-CNN. There is a performance gap between FF-CNN
and BP-CNN at the beginning of the plots. However, when the number of
labeled training data is reduced, the performance degradation of the
semi-supervised FF-CNN is not as severe as that of the BP-CNN and we see
cross-over points between these two networks in all three datasets.  For
the extreme cases, we see that semi-supervised FF-CNNs outperform
BP-CNNs by 17.1\%, 8.8\%, and 6.9\% in testing accuracy with 110, 120,
and 190 labeled data for MNIST, SVHN and CIFAR-10, respectively. The
results show that the proposed semi-supervised FF-CNNs can learn from
the unlabeled data more effectively than the corresponding BP-CNNs. 

To evaluate the effectiveness of several unlabeled data usage ideas, we
compare three different settings in Table \ref{table:accuracy_ab}. The
best results come from using selected unlabeled training data. This is
particularly obvious for the SVHN dataset. There is around 5\%
performance gain by eliminating low quality unlabeled samples. 

\subsection{Ensembles of Multiple Semi-Supervised FF-CNNs}

The performance of the proposed ensemble system by fusing all diversity
types is shown in Fig. \ref{fig:acc_2}. We see that ensembles can boost
classification accuracy by a large gain. There are about 2\%, 5\% and
8\% performance improvements against individual semi-supervised FF-CNNs,
and the ensemble results with the smallest labeled portion achieve test
accuracy of 89.6\%, 49.5\%, and 41.4\% for MNIST, SVHN and CIFAR-10,
respectively. 

We further examine the performance of different diversity types, and
show the relation between test accuracy and ensemble complexity in Fig.
\ref{fig:acc_overall}, where we test ensemble systems using different
diversity types: 1) an ensemble of four semi-supervised FF-CNNs with
varied filter sizes and filter numbers in two conv layers as listed in
Table \ref{table:set} (T1); 2) an ensemble of color input images in
different color spaces (i.e. RGB, YCbCr, and Lab), where three channels
separately for the latter two color spaces are treated separately (T2);
and 3) an ensemble of nine semi-supervised FF-CNNs obtained by taking
different input images computed from filtered greyscale images with 3x3
Laws filters \cite{pratt2007digital} (T3). 

As shown in the figure, the most efficient ensemble system among all
designs is to fuse four different semi-supervised FF-CNNs with T1
diversity which yields a performance gain of 6.4\%. In general, an
ensemble of more semi-supervised FF-CNNs provides higher testing
accuracy. The best performance achieved is 63.1\% by combining all three
diversity types. As compared with that of the single FF-CNN trained with
{\em all} labeled data, the performance of the ensemble of
semi-supervised FF-CNNs trained with 1/16 labeled data is only slightly
lower by 0.6\%. 

\section{Conclusion and future work}\label{sec:conclusion}

A semi-supervised learning framework using FF-CNNs for image
classification was proposed.  It was demonstrated by experimental
results on three benchmark datasets (MNIST, SVHN, and CIFAR-10) that the
semi-supervised FF-CNNs offer an effective solution.  The ensembles of
multiple semi-supervised FF-CNNs can boost the performance furthermore.
Two extensions of this work are under current investigation. One is
incremental learning. The other is decision fusion in the spatial
domain.  FF-CNNs provide a more convenient tool than the BP-CNN in both
contexts.  We will explore them furthermore in the near future. 

\bibliographystyle{IEEE}
\bibliography{refs}

\end{document}